\documentclass[opre,nonblindrev]{informs3}

\OneAndAHalfSpacedXI 

\usepackage{endnotes}
\usepackage{endnotes}
\usepackage{booktabs}
\usepackage{multirow}
\usepackage{longtable}
\let\footnote=\endnote
\let\enotesize=\normalsize
\def\notesname{Endnotes}%
\def\makeenmark{\hbox to1.275em{\theenmark.\enskip\hss}}
\def\enoteformat{\rightskip0pt\leftskip0pt\parindent=1.275em
  \leavevmode\llap{\makeenmark}}

\usepackage{graphicx}
\usepackage{subfigure}
\usepackage{natbib}
 \bibpunct[, ]{(}{)}{,}{a}{}{,}%
 \def\bibfont{\small}%
\usepackage{hyperref}
\hypersetup{hypertex=true,
            colorlinks=true,
            linkcolor=blue,
            anchorcolor=blue,
            citecolor=blue}
\TheoremsNumberedThrough     
\ECRepeatTheorems
\usepackage{url}
\EquationsNumberedThrough    

\usepackage{subfigure, epsfig, epstopdf}
\usepackage{graphicx,amssymb}
\usepackage{multirow, multicol}
\usepackage{amsmath,etoolbox}
\usepackage{color}
\usepackage{setspace}
\usepackage{mathrsfs}
\usepackage{dsfont}
\usepackage{bbm}
\usepackage{textcomp}

\usepackage{tikz} 
\usetikzlibrary{shapes.geometric, arrows} 
\usetikzlibrary{snakes}

\usepackage[]{algorithm}
\usepackage{algpseudocode}

\usepackage{array}
\usepackage{threeparttable}
\usepackage{graphicx}
\usepackage{verbatim}

\usepackage{multibib}
\newcites{appendix}{References}

\patchcmd{\subequations}
 {\def\theequation{\theparentequation\alph{equation}}}
 {\def\theequation{\theparentequation.\arabic{equation}}}
 {}{}

\graphicspath{{figure/}}

\begin{document}


\RUNAUTHOR{Liu et al.}

\RUNTITLE{Multi-Modal Video Feature Extraction for Popularity Prediction}

\TITLE{Multi-Modal Video Feature Extraction for Popularity Prediction}

\ARTICLEAUTHORS{%
\AUTHOR{Haixu Liu, Wenning Wang, Haoxiang Zheng, Penghao Jiang, Qirui Wang, Ruiqing Yan, Qiuzhuang Sun}
\AFF{The University of Sydney, Camperdown, New Souuth Wales, 2000, Australia, \EMAIL{hliu2490@uni.sydney.edu.au}} 
} 

\ABSTRACT{%

This work aims to predict the popularity of short videos using the videos themselves and their related features. Popularity is measured by four key engagement metrics: view count, like count, comment count, and share count.
This study employs video classification models with different architectures and training methods as backbone networks to extract video modality features. Meanwhile, the cleaned video captions are incorporated into a carefully designed prompt framework, along with the video, as input for video-to-text generation models, which generate detailed text-based video content understanding. These texts are then encoded into vectors using a pre-trained BERT model. Based on the six sets of vectors mentioned above, a neural network is trained for each of the four prediction metrics.
Moreover, the study conducts data mining and feature engineering based on the video and tabular data, constructing practical features such as the total frequency of hashtag appearances, the total frequency of mention appearances, video duration, frame count, frame rate, and total time online. Multiple machine learning models are trained, and the most stable model, XGBoost, is selected. Finally, the predictions from the neural network and XGBoost models are averaged to obtain the final result. Our team ultimately achieves first place on the leaderboard.
The code, video features extracted by our model and final result are  \href{https://colab.research.google.com/drive/1aemw6EzevBoBZmIEerC7Cq3M2ITbwniO?usp=sharing}{Google Colab}, \href{https://drive.google.com/file/d/1GU2JQBtv_UJpJTT6U5ikRqhsoYuFTRNY/view?usp=sharing}{Google Drive} and
\href{https://drive.google.com/file/d/1o_P_93GjqPgJy0DHs-IFv1AX3rlINory/view?usp=sharing}{Final Result}.





}%



\maketitle

%


\setcounter{equation}{0}
\setcounter{lemma}{0}
\setcounter{section}{0}
\setcounter{theorem}{0}
\renewcommand*{\theHequation}{\arabic{section}.\arabic{equation}} 
\renewcommand*{\theHlemma}{\arabic{section}.\arabic{lemma}} 
\renewcommand*{\theHsection}{\arabic{section}.\arabic{section}} 
\renewcommand*{\theHtheorem}{\arabic{section}.\arabic{theorem}} 


\newpage
\section{Introduction}
In recent years, short videos have become a dominant content format across various fields, especially in social media. 
Accurate popularity prediction models help content creators customize their videos, optimize content creation, and enhance viewer engagement. 
Therefore, understanding and predicting the popularity of short video content is increasingly important. This work aims to predict the popularity of short videos using the videos themselves and their related features (such as release time, number of followers of the author, video description, etc.). Popularity is measured by four key engagement metrics: view count, like count, comment count, and share count. The prediction accuracy is evaluated using mean absolute percentage error (MAPE).

\vspace{-0.3cm}
\begin{figure}[htbp]
    \centering
    \includegraphics[height=8.8cm]{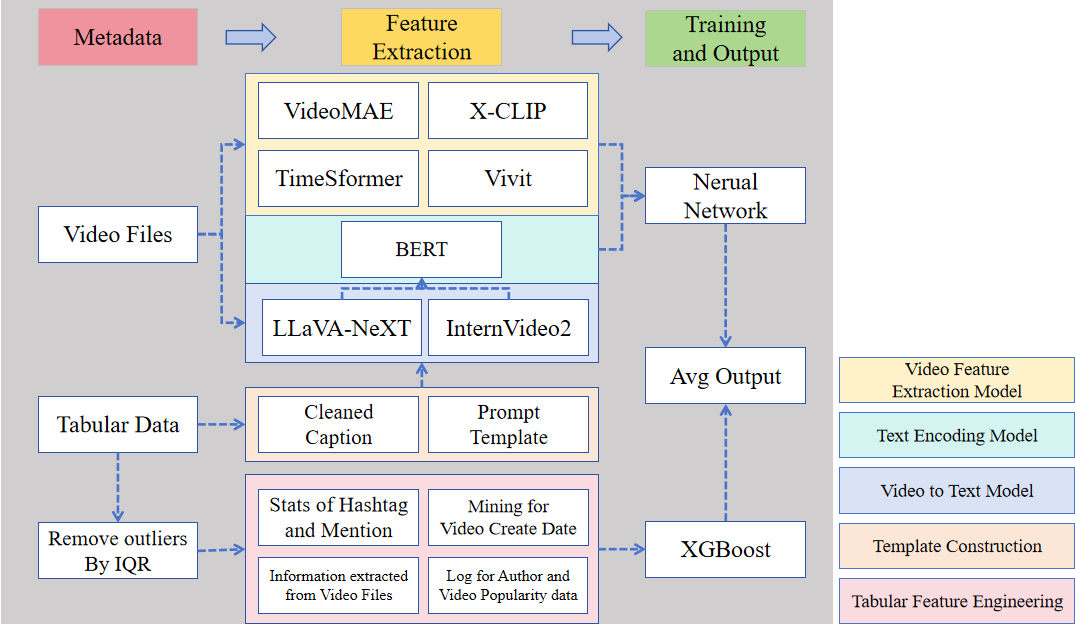 } 
    \caption{The workflow of our model.} 
    \label{Workflow} 
\end{figure}
\vspace{-0.7cm}

This work employs video classification models with different architectures and training methods, including TimeSformer \citep{bertasius2021space}, ViViT \citep{arnab2021vivit}, VideoMAE \citep{tong2022videomae}, and X-CLIP \citep{ni2022expanding}, as backbone networks to extract video modality features. 
Meanwhile, the cleaned video captions are incorporated into a carefully designed prompt framework along with the video as input for video-to-text generation models LLaVA-NeXT \citep{zhang2024llavanextvideo,liu2023llava} and InternVideo2 \citep{wang2024internvideo2}, generating detailed text-based video content understanding. These texts are then encoded into vectors using a pre-trained BERT model. Based on the six sets of vectors mentioned above, a neural network is trained for each of the four prediction metrics.

Additionally, this work conducts data mining and feature engineering based on the video and tabular data, constructing practical features such as the total frequency of hashtag (\#) appearances, the total frequency of mention (@) appearances, video duration, frame count, frame rate, and total time online. Multiple machine learning models are trained, and the most stable model, XGBoost \citep{chen2016xgboost}, is selected. Finally, the predictions from the neural network and XGBoost models are averaged to obtain the final result.

\section{Methodology}

\subsection{Tabular Feature Engineering}
Table-based data mining and feature engineering are key highlights of this work. 
The features constructed in this study significantly improve model accuracy compared to the original features.
First, the \texttt{moviepy.editor} library is used to extract information such as the duration, total frame count, frame rate, width, and height of each video. 
We assume that variations in video duration may lead to different viewing and interaction patterns, while quality metrics like frame rate and resolution might affect the viewing experience, thereby influencing the interaction rate.
However, damaged videos may fail to provide these features, so the missing values in these features are imputed. Considering that videos posted by the same author may have consistent characteristics, median imputation is applied to maintain the central tendency of the data and reduce the impact of outliers. Specifically, the missing values in both the training and test datasets are filled using the median of these features for each \texttt{author\_id} in the training data.

Next, feature engineering is performed on the \texttt{video\_create\_date} attribute to capture time-related patterns that may influence interactions. 
The timestamps are first converted into year, month, day, and hour, and it is determined whether the date falls on a U.S.\ holiday. 
Based on the hour, the time is categorized into ``working hours,'' ``leisure hours,'' or ``sleeping hours.'' The hypothesis is that the time of day when a video is posted, as well as whether it is a holiday, could influence user activity and behavior online, thereby affecting the interaction rate. Specifically, the \texttt{video\_create\_date} is a timestamp, and it is normalized so that the earliest posted video in the dataset has a value of 0, and the most recent has a value of 1. 
The intuition to construct this feature is that the longer a video has been posted, the more interactions it accumulates.

Considering that short video promotion often involves recommendation algorithms that are related to the hashtags and mentions in the video captions, the ideal approach would be to use the TikTok API to obtain hashtag usage counts and mention user counts. However, since our team is based in Australia and is restricted from using the TikTok API due to policy constraints, this study proposes an alternative feature engineering scheme. 
Specifically, regular expressions are used to extract all hashtags and mentions from the video description text, and the frequency of each hashtag and mention across all samples is calculated. For each video, the number of hashtags and mentions in its caption is counted, and the sum of all hashtag frequencies and the sum of all mention frequencies are computed, resulting in four new features. These features capture key factors in social media interactions, such as popular hashtags and mentions that may increase a video's exposure and interaction rate. Hashtags can indicate trending topics or social network relationships, thereby improving the model's ability to predict interactions.

Finally, we apply logarithmic transformations to features with large-scale differences or that are highly skewed (\texttt{author\_follower\_count}, \texttt{author\_following\_count}, \texttt{author\_total\_heart\_count}, and \texttt{author\_total\_video\_count}), as well as the target variable. 
By applying a natural logarithm plus one transformation to these features, skewness is reduced, bringing the distributions closer to normality. 
The benefit of this approach is to improve the model's robustness, avoiding instability in training due to extreme values. 
Additionally, many machine learning models perform better when features approximate a normal distribution.

\subsection{Video Feature Extraction}\label{subsec:extraction}
This work first checks the availability of training set videos, as some of the videos are corrupted to varying degrees. Thus, using the \texttt{ffmpeg} command, errors during video reading are ignored, and the healthy frames are uniformly encoded using more robust H.264 encoding. After preprocessing, only 1588 videos are retained.

Next, we select four state-of-the-art (SOTA) neural network architectures and training strategies for video feature extraction.
The videos are downsampled according to the input requirements of each network, and features are extracted using the pre-trained weights of these models. 
The four selected models are summarized below:

\begin{itemize}
    \item \textbf{TimeSformer}: A video classification model developed by Meta. 
    Its token acquisition process is similar to ViT, where a fixed number of frames are sampled from the video, and each frame is further divided into small patches. 
    These patches are linearly embedded into high-dimensional feature vectors. The Divided Space-Time Attention mechanism is used to compute temporal and spatial attention separately, replacing the self-attention in the classic Transformer.
    
    \item \textbf{ViViT}: Developed by Google, this video classification model introduces a method called Tubelet embedding based on the concept of 3D convolution. It fuses temporal and spatial information at the token stage, rather than during the encoder stage like TimeSformer. The Factorised encoder is introduced to replace the Transformer architecture, which uses separate spatial and temporal encoders.
    
    \item \textbf{VideoMAE}: Developed by Nanjing University and Tencent AI Lab, this video understanding model is based on the Masked Autoencoders architecture. 
    It is a self-supervised learning framework where parts of the input video are masked, and the model learns from the remaining parts to predict the masked sections, enhancing understanding of spatiotemporal features in the video.
    
    \item \textbf{X-CLIP}: Developed by Microsoft as an extension of OpenAI's CLIP, X-CLIP aligns video and text modalities. Its core is to learn the semantic correspondence between video and text using contrastive learning. However, it only aligns semantics and cannot generate text.
\end{itemize}

It is worth noting that the above four video feature extraction models are selected because they are \textit{pre-trained on different datasets}. 
This strategy enhances the model's performance.
We design an ablation study to verify which feature vectors are redundant in each model; see Table~\ref{key1}.

To further explore the data, this work considers how to better utilize the information in the captions to guide the model to understand the video content and shooting techniques from the user's perspective. 
After multiple adjustments, a prompt framework is constructed as shown below:

\begin{quote}
\textit{Here's the caption of the video: \{caption\}. 
Please carefully answer the following questions based on the caption and video:
\begin{enumerate}
    \item Can you analyze the video in terms of its content interestingness, including emotional appeal, engagement strategies, pacing, and uniqueness?
    \item How do the scene transitions affect pacing, engagement, and narrative flow?
    \item How do camera movements like panning, zooming, and tracking enhance storytelling, emotional tone, and immersion?
    \item What is the central plot conflict, and how does it affect character development and viewer engagement?
    \item How would you evaluate the quality of the presentation, including visuals, audio, and production techniques?
    \item What makes the video entertaining, and how do engaging moments and visual storytelling contribute to its popularity?
    \item How does the genre influence both its initial and long-term popularity?
    \item Finally, how does the emotional tone conveyed through visuals, music, and narration impact viewer connection?
\end{enumerate}}
\end{quote}

Subsequently, the prompt and video are input into LLaVA-NeXT and InternVideo2, with each model separately generating responses of 512 tokens. 
These two models work as follows.
\begin{itemize}
    \item \textbf{LLaVA-NeXT}: A video encoding model based on CLIP ViT (LLaVA) and a large language model based on Llama3-8B. A simple linear layer maps video features into the word embedding space, allowing Llama3-8B \citep{dubey2024llama} to generate natural language outputs, such as video descriptions or question answering.
    
    \item \textbf{InternVideo2}: Based on the video encoder InternVideo2 and the large language model Mistral-7B \citep{jiang2023mistral}, this video understanding model introduces a progressive training strategy. By gradually increasing the task complexity, it stabilizes and strengthens the model’s performance on video-text fusion tasks.
\end{itemize}

The outputs of the above two models are encoded into 768-dimensional vectors using BERT \citep{kenton2019bert}, which are then fed into neural network models for training to predict popularity as described in Section~\ref{subsec:train}.

\subsection{Model Training}\label{subsec:train}
We notice a few videos with very high interaction metrics. 
While this is reasonable in reality, such videos could lead to large errors during model training. 
Even applying a logarithmic transformation to the prediction metrics cannot completely eliminate its interference with the training process. Therefore, this study removes those outliers based on the interquartile range (IQR).

Next, we use the preprocessed data to train ensemble tree models for prediction. 
The hyperparameters are tuned using Bayesian optimization and ten-fold cross-validation. 
The tuning results show that the XGBoost model achieves the best MAPE in the tabular modality; see Table~\ref{key2}.
We visualize the feature importance of the XGBoost model after training; see Figures \ref{fig:Feature_importance_1}--\ref{fig:Feature_importance_4}.

We also train neural network models to fuse the extracted video features in Section~\ref{subsec:extraction}. 
Specifically, all input features are first mapped to a unified length through fully connected layers. 
For features extracted from natural language using BERT, we apply a combination of the GELU activation function and LayerNorm. 
For features extracted from the video feature extraction models, we apply a combination of ReLU and BatchNorm. 
After concatenating all vectors mapped to the unified length, the features pass through several sets of linear layers with decreasing neuron numbers and Dropout layers to make predictions.
Then, the six sets of feature vectors extracted from the six models mentioned in Section~\ref{subsec:extraction} are input into the neural network model. 
We train four neural network models, corresponding to four interaction metrics, using early stopping.

\subsection{Results}
By observing the outputs of the neural network and XGBoost, the study finds that the neural network tends to give conservative predictions, likely because its training process is based on the global optimization of MSE Loss. When the training set is right-skewed, making smaller overall predictions is more beneficial for the convergence of the Loss than making larger predictions. 
On the other hand, XGBoost is based on an ensemble of multiple decision trees, where each leaf node takes the average output. If there are samples with larger labels in a leaf node, it tends to give an exaggerated prediction.
Therefore, we average the outputs of both models and find that the MAPE of the new averaged output is lower than that of the previous models; see Table~\ref{fig:density}. Our team ultimately achieves first place on the leaderboard.

\begin{table}[ht]
\centering
\caption{Final MAPE (in \%).}
\renewcommand{\arraystretch}{0.8}
\begin{tabular}{*{9}{c}}
\toprule
& Comment & Heart & Play & Share \\
\midrule
Video & 78.45 & 83.23 & 86.50 & 85.04 \\
Tabular & 63.51 & 65.31 & 63.46 & 94.78\\
\textcolor{red}{Final Fusion} & \textcolor{red}{62.56} & \textcolor{red}{65.28} & \textcolor{red}{63.38} & \textcolor{red}{80.23}\\
\bottomrule
\end{tabular}
\end{table}

\section{Conclusion}

We summarize this study as follows.

1. In cases where there is a significant magnitude difference in the labels or a highly-skewed distribution, if the sample size cannot support highly accurate regression predictions, neural networks tend to provide conservative predictions, while ensemble tree models tend to give more expansive predictions. Averaging the results can help offset some errors.
   
2. This work proposes an effective method for constructing features based on the statistics of Hashtags and Mentions to enhance model performance. According to the feature importance chart, two extracted features—video duration and the time since the video was published—are particularly significant. Additionally, removing outliers using the IQR method can significantly improve the regression performance of tree models.

3. Among all open-source video feature extraction models, X-CLIP performs the best in predicting the popularity of short videos, while LLaVA-NeXT is clearly more suitable for generating text for short video understanding tasks.

\section{Team Members}

\begin{itemize}
\item Haixu Liu, the University of Sydney, hliu2490@uni.sydney.edu.au
\item Qiuzhuang Sun, the University of Sydney, qiuzhuang.sun@sydney.edu.au
\item Wenning Wang, the University of Sydney, wwan2926@uni.sydney.edu.au
\item Penghao Jiang, University of New South Wales, pjia0498@uni.sydney.edu.au
\item Haoxiang Zheng, University of New South Wales, z5574306@ad.unsw.edu.au
\item Qirui Wang, University of New South Wales, qirui.wang@unilabs.org.cn
\item Ruiqing Yan, University of New South Wales, ruiqing.yan@unilabs.org.cn
\end{itemize}
\newpage
\bibliographystyle{informs2014} 
\bibliography{ref} 
\newpage
\section*{Appendix}
\begin{APPENDICES}

{\begin{table}[ht]
\centering
\caption{A summary of prediction results.}
\label{key2}
\renewcommand{\arraystretch}{0.8}
\begin{tabular}{*{9}{c}}
\toprule
\multirow{2}*{Model} & \multicolumn{2}{c}{Comment} & \multicolumn{2}{c}{Heart} & \multicolumn{2}{c}{Play} & \multicolumn{2}{c}{Share} \\
\cmidrule(lr){2-3} \cmidrule(lr){4-5} \cmidrule(lr){6-7} \cmidrule(lr){8-9}
  & MSE & MAPE & MSE & MAPE & MSE & MAPE & MSE & MAPE  \\ 
\midrule
CatBoost & 2.55E+06 & 0.68 & 3.10E+10 & 0.76 & 3.55E+12 & 0.71 & 3.81E+07 & 1.07 \\
\textcolor{red}{XGBoost} & \textcolor{red}{2.53E+06} & \textcolor{red}{0.64} & \textcolor{red}{3.10E+10} & \textcolor{red}{0.65} & \textcolor{red}{3.53E+12} & \textcolor{red}{0.63} & \textcolor{red}{3.80E+07} & \textcolor{red}{0.95} \\
LightGBM & 2.51E+06 & 0.69 & 3.07E+10 & 0.74 & 3.55E+12 & 0.75 & 3.80E+07 & 1.10 \\
LCE Model & 2.51E+06 & 0.66 & 3.11E+10 & 0.76 & 3.56E+12 & 0.73 & 3.81E+07 & 1.03 \\
LinearRegression & 2.59E+06 & 0.80 & 3.10E+10 & 0.93 & 3.56E+12 & 0.88 & 3.81E+07 & 1.23 \\
Ridge & 2.58E+06 & 0.81 & 3.10E+10 & 0.95 & 3.57E+12 & 0.88 & 3.81E+07 & 1.24 \\
Lasso & 2.74E+06 & 1.33 & 3.42E+10 & 2.45 & 3.72E+12 & 1.76 & 3.83E+07 & 1.97 \\
ElasticNet & 2.74E+06 & 1.33 & 3.42E+10 & 2.45 & 3.72E+12 & 1.76 & 3.83E+07 & 1.97 \\
BayesianRidge & 2.59E+06 & 0.81 & 3.11E+10 & 0.97 & 3.58E+12 & 0.89 & 3.81E+07 & 1.26 \\
HuberRegressor & 2.58E+06 & 0.79 & 3.08E+10 & 1.02 & 3.56E+12 & 1.03 & 3.80E+07 & 1.24 \\
ARDRegression & 2.59E+06 & 0.81 & 3.10E+10 & 0.94 & 3.57E+12 & 0.88 & 3.81E+07 & 1.25 \\
SGDRegressor & 2.60E+06 & 0.83 & 3.14E+10 & 0.97 & 3.57E+12 & 0.93 & 3.81E+07 & 1.29 \\
PassiveAggressiveRegressor & 2.56E+06 & 0.88 & 3.38E+10 & 0.80 & 3.62E+12 & 0.81 & 3.70E+07 & 6.02 \\
RandomForestRegressor & 2.51E+06 & 0.68 & 3.09E+10 & 0.72 & 3.53E+12 & 0.69 & 3.81E+07 & 1.01 \\
GradientBoostingRegressor & 2.53E+06 & 0.71 & 3.12E+10 & 0.79 & 3.56E+12 & 0.77 & 3.81E+07 & 1.08 \\
AdaBoostRegressor & 2.61E+06 & 0.86 & 3.21E+10 & 0.97 & 3.61E+12 & 0.98 & 3.82E+07 & 1.20 \\
ExtraTreesRegressor & 2.47E+06 & 0.74 & 3.05E+10 & 0.75 & 3.50E+12 & 0.82 & 3.80E+07 & 1.03 \\
BaggingRegressor & 2.49E+06 & 0.71 & 3.07E+10 & 0.74 & 3.53E+12 & 0.81 & 3.81E+07 & 1.05 \\
SVR & 2.54E+06 & 0.72 & 3.08E+10 & 0.92 & 3.55E+12 & 0.89 & 3.81E+07 & 1.21 \\
NuSVR & 2.56E+06 & 0.73 & 3.12E+10 & 0.91 & 3.56E+12 & 0.88 & 3.81E+07 & 1.17 \\
LinearSVR & 2.58E+06 & 0.79 & 3.09E+10 & 1.05 & 3.57E+12 & 1.06 & 3.80E+07 & 1.28 \\
KNeighborsRegressor & 2.53E+06 & 0.76 & 3.10E+10 & 0.82 & 3.54E+12 & 0.77 & 3.80E+07 & 1.16 \\
DecisionTreeRegressor & 2.51E+06 & 0.84 & 3.07E+10 & 1.02 & 3.46E+12 & 0.98 & 3.80E+07 & 1.71 \\
MLPRegressor & 2.52E+06 & 0.73 & 3.09E+10 & 0.82 & 3.57E+12 & 0.81 & 3.81E+07 & 1.12 \\
PLSRegression & 2.57E+06 & 0.85 & 3.12E+10 & 1.00 & 3.58E+12 & 0.90 & 3.81E+07 & 1.28 \\
TabNet & 2.66E+06 & 1.43 & 3.33E+10 & 2.02 & 3.67E+12 & 1.52 & 3.82E+07 & 1.58 \\
\bottomrule
\end{tabular}
\end{table}}

\newpage
\renewcommand{\arraystretch}{0.8}
\begin{longtable}{ccccc}
    \caption{Performance metrics in terms of MAPE.}
    \label{key1}\\
    \toprule
    Feature & Share & Heart & Comment & Play \\
    \midrule
    \endfirsthead
    \multicolumn{5}{r}{}\\
    \toprule
    Feature & Share & Heart & Comment & Play \\
    \midrule
    \endhead
    \bottomrule
    \multicolumn{5}{c}{}\\
    \endfoot
    \bottomrule
    \endlastfoot
        1 (VideoMAE-Base) & 89.04 & 91.04 & 89.10 & 93.54 \\
        2 (ViViT-B 16x2) & 92.45 & 91.72 & 85.76 & 85.96 \\
        3 (TimeSformer) & 91.04 & 88.24 & 82.87 & 87.68 \\
        4 (X-CLIP Base Patch32) & 85.46 & 81.45 & 81.06 & 81.77 \\
        5 (LLaVA NeXT Video Model 7B hf) & 89.33 & 88.48 & 80.19 & 88.14 \\
        6 (InternVideo2) & 92.19 & 88.79 & 82.87 & 92.82 \\
        1+5   & 90.18 & 90.81 & 82.42 & 88.21 \\
        2+5   & 89.45 & 87.37 & 80.17 & 84.23 \\
        3+5   & 86.78 & 89.22 & 78.58 & 84.82 \\
        4+5   & 84.70 & 85.66 & 81.37 & 81.39 \\
        1+6   & 91.08 & 92.59 & 89.52 & 88.15 \\
        2+6   & 87.47 & 89.83 & 85.82 & 81.98 \\
        3+6   & 90.03 & 91.01 & 79.79 & 82.10 \\
        4+6   & 87.53 & 88.99 & 78.96 & 82.84 \\
        1+2+3+4 & 85.25 & 83.66 & 78.80 & 84.31 \\
        5+6   & 88.70 & 92.96 & 81.20 & 89.03 \\
        2+3+4+5+6& 86.73 & 82.13 & 81.47 & 87.09 \\
        1+3+4+5+6& 88.90 & 87.83 & 82.31 & 83.73 \\
        1+2+4+5+6& 86.74 & 83.19 & 83.36 & 83.46 \\
        1+2+3+5+6& 82.55 & 86.41 & 79.52 & 83.15 \\
        1+2+3+4+6& 80.73 & 80.36 & 80.46 & \textcolor{red}{76.11} \\
        1+2+3+4+5& 79.80 & 82.63 & 80.17 & 76.18 \\
        2+3+4+6 & 86.62 & 81.38 & \textcolor{red}{73.04} & 82.85 \\
        1+3+4+6 & 86.58 & 87.21 & 79.71 & 87.08 \\
        1+2+4+6 & 85.77 & 84.64 & 77.90 & 79.52 \\
        1+2+3+6 & 89.75 & 87.44 & 79.27 & 85.55 \\
        2+3+4+5 & 87.49 & 87.83 & 76.05 & 79.13 \\
        1+3+4+5 & 85.24 & 84.60 & 74.66 & 84.28 \\
        1+2+4+5 & 87.40 & 83.65 & 86.15 & 78.97 \\
        1+2+3+5 & 88.05 & 88.24 & 78.27 & 85.65 \\
        1+2   & 89.39 & 88.23 & 82.47 & 85.06 \\
        1+3   & 82.57 & 87.07 & 79.97 & 82.39 \\
        1+4   & 82.83 & 83.01 & 77.72 & 83.95 \\
        2+3   & 87.20 & 85.93 & 78.05 & 84.34 \\
        2+4   & 80.07 & 80.93 & 77.91 & 81.75 \\
        3+4   & \textcolor{red}{78.92} & \textcolor{red}{79.55} & 76.58 & 81.88 \\
        1+2+3  & 87.55 & 90.68 & 82.90 & 86.03 \\
        1+2+4  & 88.91 & 85.26 & 81.80 & 83.77 \\
        1+3+4  & 87.60 & 85.80 & 81.16 & 79.57 \\
        2+3+4  & 88.24 & 87.03 & 76.60 & 78.73 \\
        1+2+5  & 91.09 & 90.16 & 81.99 & 82.60 \\
        1+3+5  & 91.90 & 93.70 & 85.96 & 86.25 \\
        1+4+5  & 85.17 & 89.42 & 81.73 & 83.13 \\
        2+3+5  & 90.48 & 90.32 & 79.46 & 84.76 \\
        2+4+5  & 87.58 & 86.54 & 77.59 & 86.07 \\
        3+4+5  & 86.87 & 90.41 & 77.00 & 85.62 \\
        1+2+6  & 89.21 & 91.44 & 82.56 & 85.13 \\
        1+3+6  & 90.01 & 91.02 & 86.96 & 86.49 \\
        1+4+6  & 84.09 & 89.45 & 85.59 & 83.86 \\
        2+3+6  & 90.69 & 90.45 & 82.04 & 82.96 \\
        2+4+6  & 85.58 & 83.72 & 77.77 & 78.08 \\
        3+4+6  & 88.60 & 86.59 & 73.96 & 81.34 \\
        1+2+3+4+5+6 & 85.04 & 83.23 & 78.45 & 86.50 \\
\end{longtable}

\newpage
\begin{figure}[htbp]
    \centering
    \includegraphics[width=1\textwidth]{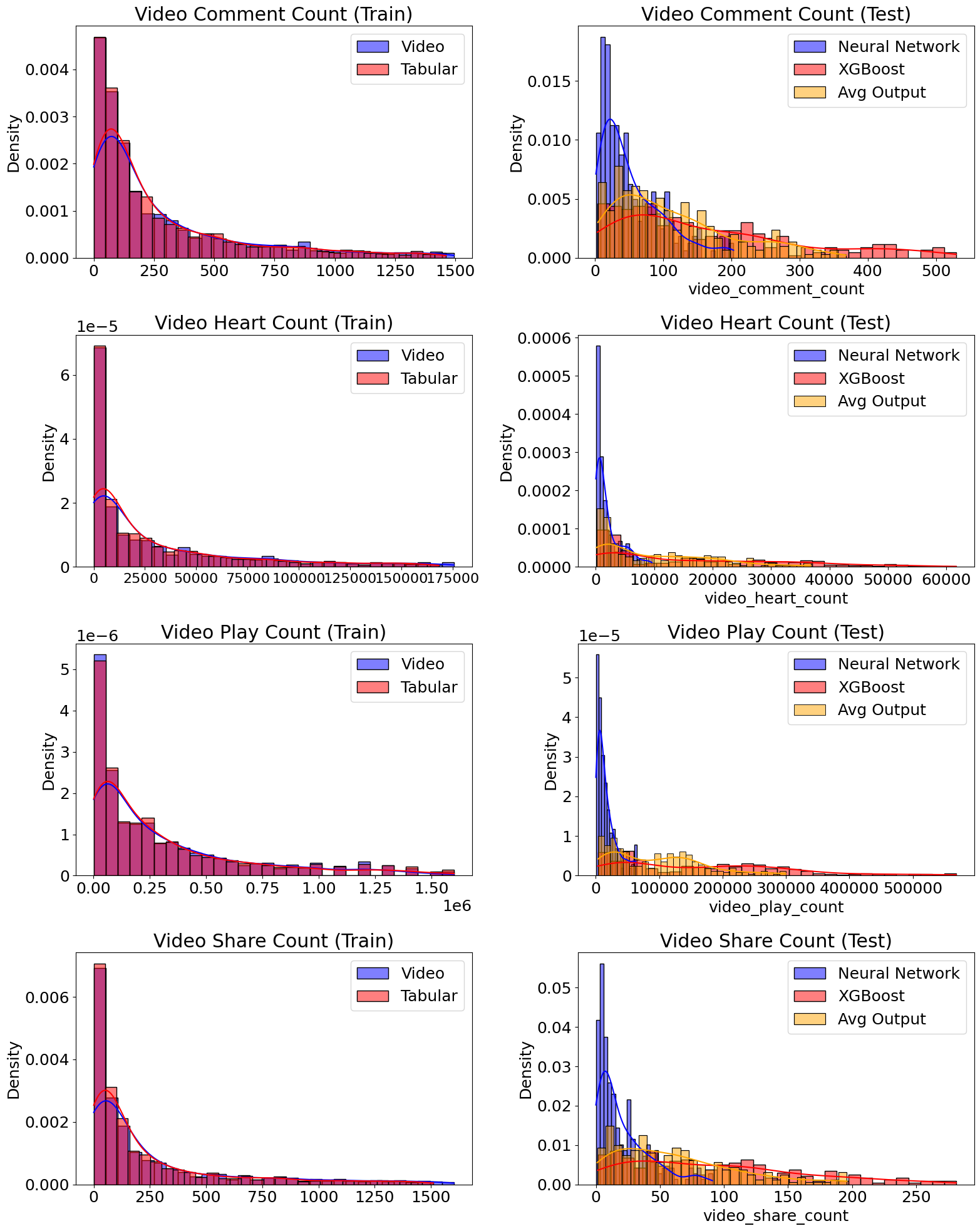 } 
    \caption{Comparison of the distributions of the outputs of the training and test sets of neural networks and XGBoost} 
    \label{fig:density} 
\end{figure}

\begin{figure}[htbp]
    \centering
    \includegraphics[height=10.8cm]{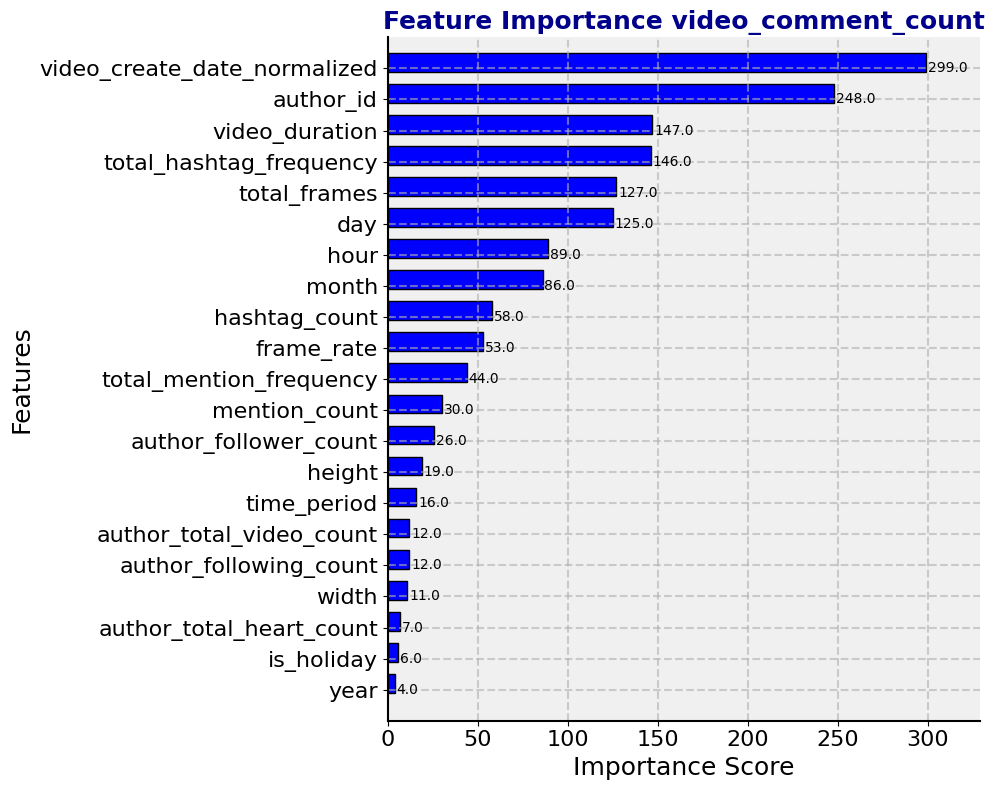} 
    \vspace{-0.5cm} 
    \caption{Feature Importance of comment prediction} 
    \label{fig:Feature_importance_1} 
\end{figure}
\vspace{-1cm} 
\begin{figure}[htbp]
    \centering
    \includegraphics[height=10.8cm]{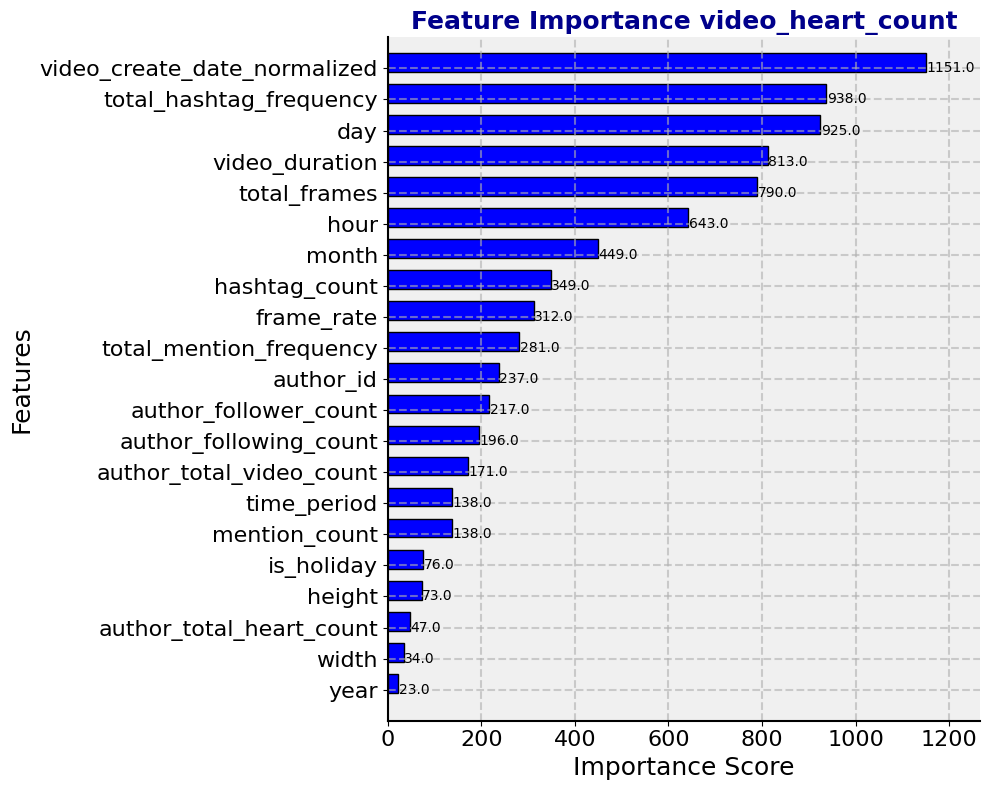} 
    \vspace{-0.5cm} 
    \caption{Feature Importanceof of heart prediction } 
    \label{fig:Feature_importance_2} 
\end{figure}
\vfill
\vspace{-1cm} 
\vfill
\begin{figure}[htbp]
    \centering
    \includegraphics[height=10.8cm]{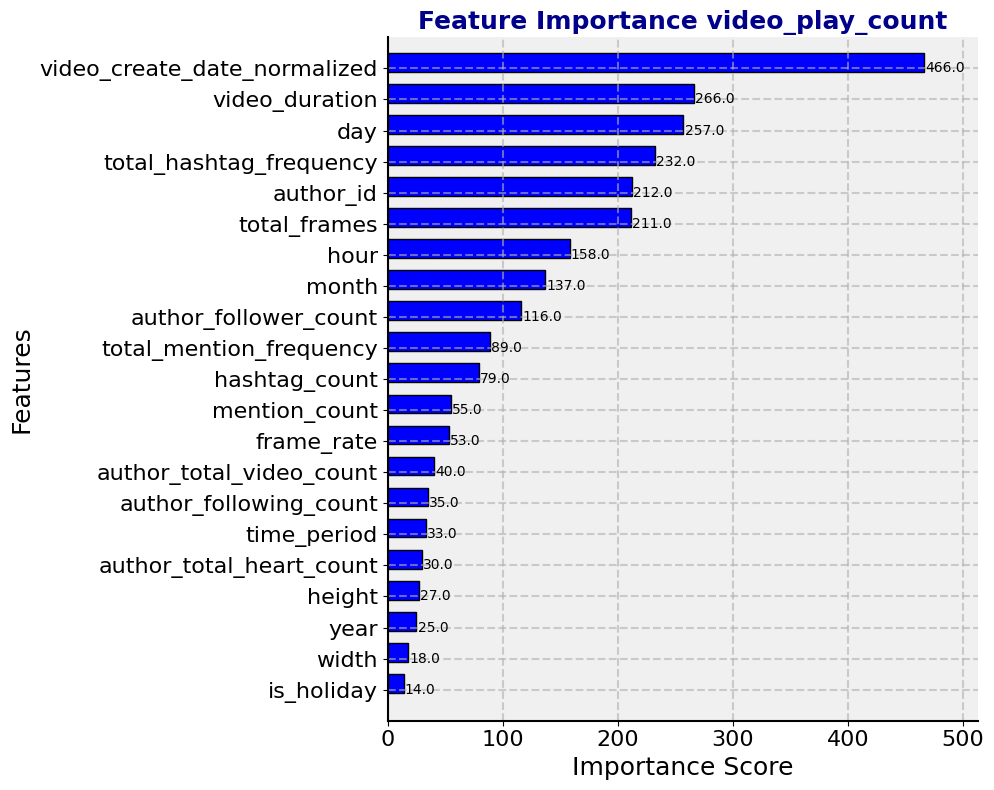} 
    \vspace{-0.5cm} 
    \caption{Feature Importance of play prediction} 
    \label{fig:Feature_importance_3} 
\end{figure}
\vfill
\vspace{-1cm} 
\vfill
\begin{figure}[htbp]
    \centering
    \includegraphics[height=10.8cm]{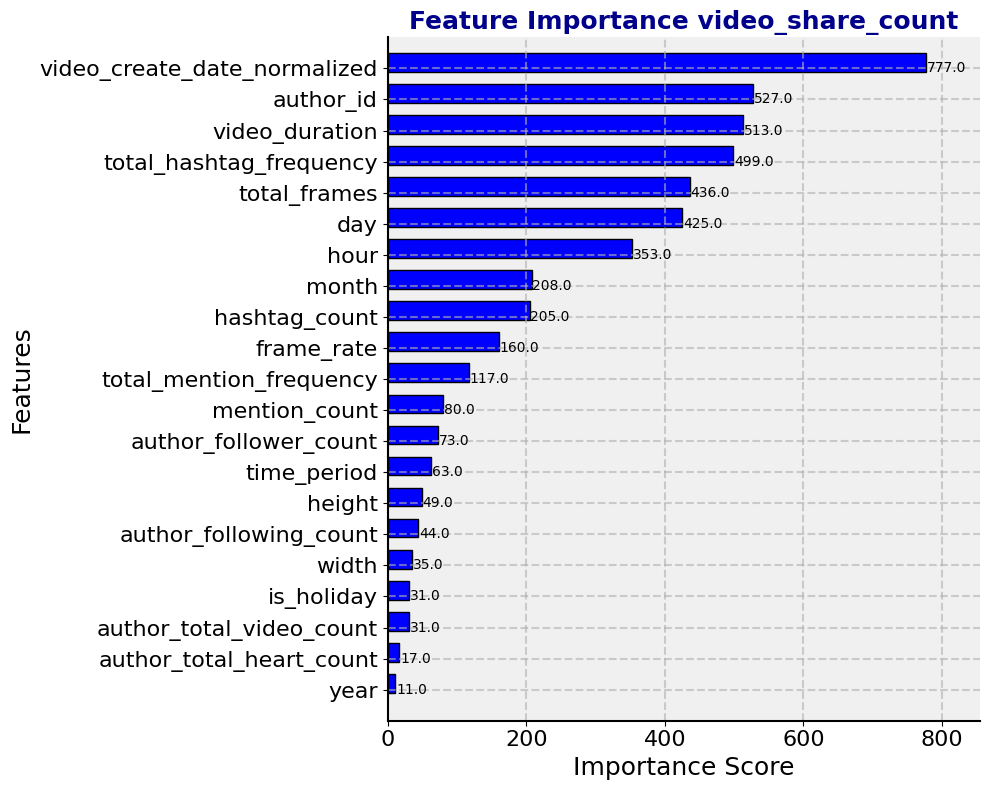} 
    \vspace{-0.5cm} 
    \caption{Feature Importance of share prediction } 
    \label{fig:Feature_importance_4} 
\end{figure}

\end{APPENDICES}

\end{document}